\documentclass[pdflatex, sn-basic]{sn-jnl}% Basic Springer Nature Reference Style/Chemistry Reference Style

\usepackage{graphicx}%
\usepackage{multirow}%
\usepackage{amsmath,amssymb,amsfonts,bm}%
\usepackage{amsthm}%
\usepackage{mathrsfs}%
\usepackage[title]{appendix}%
\usepackage{xcolor}%
\usepackage{textcomp}%
\usepackage{manyfoot}%
\usepackage{booktabs}%
\usepackage{algorithm}%
\usepackage{algorithmicx}%
\usepackage{algpseudocode}%
\usepackage{listings}%
\usepackage{natbib}
\usepackage{subcaption}
\usepackage{makecell}
\usepackage{orcidlink}

\DeclareMathOperator*{\argmin}{argmin}

\begin{document}

\title[Article Title]{Constraint Guided AutoEncoders for Joint Optimization of Condition Indicator Estimation and Anomaly Detection in Machine Condition Monitoring}

\author*[1,2]{Maarten Meire \orcidlink{0000-0001-7317-6523}}\email{maarten.meire@kuleuven.be}
\equalcont{These authors contributed equally to this work.}

\author[1,2]{Quinten Van Baelen\orcidlink{0000-0003-2863-4227}}\email{quinten.vanbaelen@kuleuven.be}
\equalcont{These authors contributed equally to this work.}

\author[3]{Ted Ooijevaar\orcidlink{0000-0001-7120-7832}}
\email{ted.ooijevaar@flandersmake.be}

\author[1,2]{Peter Karsmakers\orcidlink{0000-0001-8119-6823}}\email{peter.karsmakers@kuleuven.be}

\affil[1]{KU Leuven, Dept. of Computer Science; Leuven.AI,  3000 Leuven, Belgium}

\affil[2]{Flanders Make @ KU Leuven}

\affil[3]{Flanders Make vzw, CoreLab MotionS, 3001 Leuven, Belgium}

\abstract{The main goal of machine condition monitoring is, as the name implies, to monitor the condition of industrial applications. 
The objective of this monitoring can be mainly split into two problems.
A diagnostic problem, where normal data should be distinguished from anomalous data, otherwise called Anomaly Detection (AD), or a prognostic problem, where the aim is to predict the evolution of a Condition Indicator (CI) that reflects the condition of an asset throughout its life time.
When considering machine condition monitoring, it is expected that this CI shows a monotonic behavior, as the condition of a machine gradually degrades over time.
This work proposes an extension to Constraint Guided AutoEncoders (CGAE), which is a robust AD method, that enables building a single model that can be used for both AD and CI estimation. For the purpose of improved CI estimation the extension incorporates a constraint that enforces the model to have monotonically increasing CI predictions over time.
Experimental results indicate that the proposed algorithm performs similar, or slightly better, than CGAE, with regards to AD, while improving the monotonic behavior of the CI.}
%150-250 woorden

\keywords{anomaly detection, condition estimation, autoencoders, deep learning}

\maketitle

\section{Introduction}\label{sec:introduction}

The main goal of machine condition monitoring is, as the name implies, to monitor the condition of industrial applications, and one of the crucial parts in these applications is rotating machinery.
When considering this type of machinery, it has been determined that the main cause of failure is attributed to Rolling Element Bearings (REB) \citep{Nabhan2015}.
Currently, the most common method to monitor these REB is currently by capturing and analyzing the vibrations produced by REB \citep{Hoang2019}.
However, acoustic signals have also been used as a promising alternative \citep{Pacheco2022}, and allow for easier data collection.
The work regarding this condition monitoring can be mainly split into two branches.
The first branch considers a diagnostic problem, where a signal is either classified as normal or anomalous, otherwise called Anomaly Detection (AD), or it can be further categorized into specific fault types, depending on the problem and considered method \citep{Jiang2019}. 
The other branch considers a prognostic problem, where a signal is given a condition score, or Condition Indicator (CI), that assesses the condition of a machine \citep{She2020}. 
This CI can then be used as a basis to predict the Remaining Useful Life (RUL) of a machine (component) or can be used to determine if the state of the machine is considered normal or anomalous based on some predefined threshold.
For both AD and CI estimation research has been done using signal processing, e.g. traditional clustering \citep{Knorr2000} or shallow Machine Learning (ML) \citep{Achmad2007}, and data-driven, mainly Deep Learning (DL) \citep{Ionescu2017, Said2020}, methods, with the latter receiving more attention in recent years \citep{Zhou2022, Zhang2020}.
This work will focus on using data-driven DL methods.

AD is an important task in a broad range of domains, such as, network intrusion detection \citep{Yang2022}, finance \citep{Ahmed2016}, and machine condition monitoring \citep{Kamat2020}.
The most common DL approach used for AD is an AutoEncoder (AE) or an AE-based variant \citep{Chalapathy2019}.
These methods work by encoding data into a compact latent representation in a manner that allows the original data to be reconstructed as well as possible \citep{Hinton2006}.
By applying this on normal data, it is expected that the reconstruction of anomalous data is worse, as there are differences between both types of data.
This can be seen in \citep{Oh2018}, where an AE was used to perform AD on acoustic data from a surface-mount device assembly machine.
A similar methodology was used in \citep{Givnan2022}, where a stacked autoencoder was used to perform AD on sensor data from industrial motors.
A downside of this type of approach is the inability to use anomalous data during training, which can provide valuable information, hence, approaches that allow the use of anomalous data are of interest \citep{Primus2020, Chenyu2020}.  
This additional information can be used by either semi- or fully supervised methods.
However, it is indicated in \citep{Nari2017} that fully supervised methods only have a better performance in case all types of anomalies are known during training.
A similar conclusion is made in \citep{Shen2021}, where it was indicated that incorrect labeling reduces the performance of a supervised method, while keeping these label uncertainties unlabeled in a semi-supervised approach mitigates this issue.
Since it is unlikely that all types of anomalies are known during training and label uncertainties are possible, semi-supervised methods are likely to, in general, outperform fully supervised methods. 
A commonly used semi-supervised method is Deep Support Vector Data Description (DSVDD) \citep{Ruff2018, Ruff2019}, which attempts to map normal data close to a center and abnormal data further away from the same center, and is an adaptation of the original Support Vector Data Description (SVDD) method \citep{Tax2004} into the DL framework.
However, this adaptation requires to restrict the DL network used by the method, so that trivial solutions are avoided \citep{Ruff2018, Ruff2019}: (i) the center used by DSVDD cannot be initialised as 0, (ii) the network cannot use biases, and (iii) the network cannot have bounded activation functions.
In \citep{Liu2021} a DSVDD model was used to perform AD on vibration data from helicopters. They made a comparison between the data-driven DSVDD method and a traditional SVDD method with engineered features, showing an improved performance for the DSVDD method.
A similar approach was used in \citep{Peng2023} to monitor the condition of wind turbines.
The DSVDD method was also compared to traditional SVDD as well as to a regular AE, again showing an improved performance, although the difference with the AE was not large.

These restrictions were alleviated in a combined AE and DSVDD approach, where the method will still learn to map normal data close to a center, as per the DSVDD objective.
However, a trivial solution, where all the normal data is mapped to a single point, will be prevented by the added reconstruction objective of the AE \citep{Zhang2021, Zhou2021, Hojjati2024}.

As discussed earlier, next to the diagnostic AD problem, a prognostic CI estimation problem can also be considered.
DL methods have already been applied to this problem in previous works.
In \citep{Su2020} a Variational AutoEncoder (VAE) is combined with different types of Recurrent Neural Networks (RNN) to estimate the RUL of aircraft engines.
A pretrained VGG16 model was used in combination with SVDD in \citep{Mao2020}, by first adapting the VGG16 model to the bearing data, after which the features obtained from this model were passed to the SVDD algorithm to estimate the CI.
A Multi-Task DSVDD approach was used in \citep{Huadong2021} to estimate the CI for different bearings.
This was done by creating a hypersphere around a center for each different bearing and introducing a hypersphere similarity loss that attempts to overlap these hyperspheres as much as possible.
A similar approach was used in \citep{Kou2022}. However, instead of attempting to create overlapping hyperspheres, a set of centers is chosen using prototype clustering and these are used to determine the probability of a point to be assigned to one of these center.
In \citep{Wang2023} a combination of a CNN and RNN was used to perform CI estimation based on the vibration data and an enhanced empirical mode decomposition algorithm.

There are also methods that detect anomalies when the CI goes above some predefined threshold (assuming a low CI corresponds to normal behavior while a high CI corresponds to deviating behavior) \citep{Chenyu2020}.
However, determining this threshold is a nontrivial problem as it often depends on the specific application, e.g. false negatives might be undesirable, which evidently influences the threshold. Different methods to set this threshold exist. 
They can be split into three groups: (i) methods that rely on statistics from the training data, such as the mean and standard deviation or the maximum \citep{Wang2023,Kou2022}, (ii) methods that use additional data, e.g. the test set, to set a threshold to optimize a metric \citep{Garg2021}, and (iii) methods that dynamically determine the threshold based on the data that is being evaluated \citep{Jia2022}.
In this work, it is opted to focus on methods belonging to the first group, as these do not require additional data or analysis. 
It has been empirically observed that choosing a threshold based on these methods, as is expected, shows a strong dependency on the training and initialization of a model, especially when DL models are used.
A more robust thresholding method, called Constraint-Guided AutoEncoders (CGAE) was proposed in \citep{Meire2023}, where a fixed threshold was defined and constraints were applied to the training so that an encoder is enforced to map normal and anomalous data to the correct side of this threshold.
It was indicated that this method of training with constraints resulted in a robust threshold that is less dependent of the distribution of the training data in the latent space.

The aim of CGAE is to separate normal and anomalous data based on a fixed, predefined threshold. 
However, when considering machine condition monitoring as a prognostic problem, it is expected that this CI shows a monotonically increasing behavior, which is not directly linked to the objective of CGAE.
Previous works have already applied different methods to learn or enforce this monotonic behavior.
In \citep{Wu2019} 4 different monotonic curves, that emulate an expected RUL curve, were used as target for a learning objective.
A similar approach was used in \citep{Hu2020}, where features were selected based on both a correlation metric between the feature and the running time of an asset and a monotonicity metric.
The selected features were then used to construct a single CI with the normalized life of an asset as target.
While these methods allow for the construction of a monotonic CI, they require a target or label that matches the degradation pattern of the bearing, which might not be trivial, as each bearing can show a different degradation pattern \citep{Sikorska2011}.
A two step approach was used in \citep{Qi2024}, where a CI was selected by first splitting a raw vibration signal into multiple frequency bands, then calculating multiple CIs per frequency band, and finally selecting the CI that attains the highest score based on the monotonicity, trendability, and prognosability.
This CI is then provided to a SVDD model to perform AD, and when an anomaly is detected the RUL is estimated using a statistical model.
A two step approach was also used in \citep{Nieves2022} to first estimate a CI using a stochastic gradient descent regressor, and second, optimize an isotonic regression to fit this CI onto a monotonic curve.
As these methods use two steps, a change in one of the steps results in the other step needing to be reoptimized, based on that change.

This work proposes an extension to CGAE, which will be called Monotonically Constraint Guided AutoEncoder (MCGAE), where an additional monotonicity constraint is included to enforce the monotonic behavior of the CI estimations over time.
In this way the proposed method causes an AE to learn to both map normal and anomalous data to the correct side of some threshold, as per the CGAE, objective, as well as to map data from earlier in the life time of a bearing closer to the origin than data from later in the life time of a bearing. 
Since the CI is defined as the distance of the latent representation of the current measurement to the latent space origin the latter will cause the desired monotonic behavior.
Using this approach, our proposed method does not require a target that matches the expected degradation pattern of a bearing, as the monotonicity is enforced by the constraint.
The optimization of both the AD and CI estimation is also integrated into a joint optimization, meaning that a change in either will automatically also optimize the other.

The main contributions of this work are:
\begin{itemize}
    \item An extension of the CGAE algorithm that retains the ability to separate normal and anomalous data, while having an improved underlying CI by enforcing it to have a monotonic trend.
    \item A joint modelling of CI estimation and AD which allows the use of both labeled and unlabeled data, without needing to assume that the latter can be considered (mostly) as normal data.
    \item A comparison of the proposed method with existing methods using the data from two sets of life time tests of bearings: one where the health is monitored via an accelerometer and another where a microphone was used. 
\end{itemize}

The rest of this paper is structured as follows.
The methods that will be used for comparison are described in Section \ref{sec:methods}.
This is followed by detailed discussion of the proposed MCGAE method in Section \ref{sec:MCGAE}.
Then, the experimental setup, including the used datasets, preprocessing, model architectures, and evaluation metrics are explained in Section \ref{sec:setup}.
The performed experiments and corresponding results are discussed in Section \ref{sec:experimentsResults}.
An ablation study is presented in Section \ref{sec:ablation}.
Finally, a conclusion to this work is given in Section \ref{sec:conclusion} as well as possible directions for future work.

\section{Methods}\label{sec:methods}
This section will introduce the methods that will be used for comparison with the proposed MCGAE method, more specifically, AutoEncoder Deep Support Vector Data Description (AE-DSVDD) and Constraint Guided AutoEncoders (CGAE).
The data used in this work consists of normal, unlabeled, and anomalous points.
The normal points correspond to measurements taken when the machine was in a healthy condition, the unlabeled points correspond to a situation where the machine condition is considered to be unknown, and anomalous points correspond to faulty machine conditions, this will be explained in more detail in Section \ref{ssec:datasets}.

\subsection{AutoEncoder Deep Support Vector Data Description}\label{ssec:ae-dsvdd}
The learning objectives of both AE and DSVDD are combined in AutoEncoder Deep Support Vector Data Description \citep{Zhang2021}. 
The resulting objective learns to both reconstruct normal points as good as possible, as per the AE objective, and also to map the normal points as close as possible to a center point $\bm{c} \in \mathbb{R}^{D_l}$, with $D_l$ the amount of dimensions in a latent space, and anomalous points as far away as possible from $\bm{c}$. %, according to the DSVDD objective.}
This is done by learning both the weights of an encoder $\mathcal{E}(\cdot | \theta_\mathcal{E})$, where $\theta_\mathcal{E}$ are the weights of the encoder of an AE, and those of a decoder $\mathcal{D}(\cdot | \theta_\mathcal{D})$, where $\theta_\mathcal{D}$ are the weights of the decoder of an AE, by solving the following objective

\begin{align}\label{eq:AE-DSVDD}
    \argmin_{\theta_\mathcal{E},\theta_\mathcal{D}}&\quad\frac{1}{N}\sum\limits_{\bm{x} \in \mathcal{N}}\left(\left\|\bm{x}-\mathcal{D}\left( \mathcal{E}\left(\bm{x} \mid \theta_\mathcal{E}\right) \mid \theta_\mathcal{D}\right)\right\|^2  + \left\|\mathcal{E}(\bm{x}|\theta_\mathcal{E})-\bm{c}\right\|^2\right) \\ &\quad\quad+ \lambda\frac{1}{A}\sum\limits_{\bm{x} \in \mathcal{A}}\left(\left\|\mathcal{E}(\bm{x}|\theta_\mathcal{E})-\bm{c}\right\|^2\right)^{-1},\nonumber
\end{align}

where $N$ and $A$ are the amount of normal and anomalous points, respectively, $\mathcal{N}=\{n_i\}_{i=1}^{N}$ is the set of normal points, $\mathcal{A}=\{a_i\}_{i=1}^{A}$ is the set of anomalous points, $\lambda > 0$ is a hyperparameter that balances the weight of anomalous data in comparison with normal data, and $||\cdot||$ is the $L_2$-norm.
Do note that the objective in \citep{Zhang2021} only uses normal points, whereas \eqref{eq:AE-DSVDD} also incorporates anomalous data.
This addition of anomalous data has already been used in \citep{Meire2023}.

Determining the CI for an input point $\bm{x}$ can then be done by using the distance of the encoding of $\bm{x}$ to the center $\bm{c}$ as follows

\begin{equation}\label{eq:AE-DSVDD_CI}
    f_{CI}(\bm{x}) = \left\|\mathcal{E}(\bm{x}|\theta_\mathcal{E})-\bm{c}\right\|^2. 
\end{equation}

This CI can then be used to determine if $\bm{x}$ should be considered as normal or anomalous by comparing it to a threshold $T$ using the following classification rule

\begin{equation}\label{eq:AE-DSVDD_AD}
    \hat{y}(\bm{x})=
    \begin{cases}
      1, & \ f_{CI}(\bm{x}) > T \\
      -1, & \ f_{CI}(\bm{x}) \leq T,
    \end{cases}
\end{equation}
where $\hat{y}=-1$ and $\hat{y}=1$ corresponds to a normal and anomalous point, respectively.

As already mentioned in Section \ref{sec:introduction}, there are different methods to determine $T$, and in this work three methods, that were described in \citep{Meire2023}, will be used.
The first method determines an "optimal" threshold $T_{opt}$. The value is chosen such that a chosen metric is optimized for the considered test set.
As this is the best possible threshold, the performance can be regarded as an upper bound.
The second method determines a threshold $T_{train}$ by computing the mean $\mu_n$ and standard deviation $\sigma_n$ of the estimated CI for the normal training data and then sets $T_{train}$ to $\mu_n + 3\sigma_n$.
This is one of the more simple and basic methods to determine a threshold.
The third and final method attempts to fit a sigmoid on the estimated CI of the normal training data, such that the median and $99$-th percentile are receiving values $0.25$ and $0.5$ at the sigmoid's output, respectively.
The supremum, the maximum value of the sigmoid, is set to 1.
By setting the threshold to $0.6$ on this sigmoid and then inverting the function, the value of this threshold $T_{sigmoid}$ can be determined.
Note that the latter has similarities with the output layer of DL models as they typically employ sigmoid activation functions. %on the output layer of a DL model.}

\subsection{Constraint Guided Autoencoder}

In the Constraint Guided AutoEncoder (CGAE) \citep{Meire2023} the learning objective of an AE is subjected to a set of constraints: (i) normal data should be mapped inside of a sphere, with radius $R_1$, around the origin, and (ii) anomalous data should be mapped outside of a larger sphere, with radius $R_2$, around the origin.
This learning objective is made with the assumption that at least some anomalous data is available, so that the model can learn to distinguish between normal and anomalous data, and it is also assumed that $R_1 + \delta < R_2$ with $\delta$ some positive constant, so that there is a separation between the mapping of the normal and anomalous data, which is expected to lead to a better generalization.
The resulting constrained learning objective is as follows

\begin{align}\label{eq:CGAE}
     \argmin_{\theta_\mathcal{E},\theta_\mathcal{D}}&\quad\frac{1}{N}\sum\limits_{\bm{x} \in \mathcal{N}}\left\|\bm{x}-\mathcal{D}\left( \mathcal{E}\left(\bm{x} \mid \theta_\mathcal{E}\right) \mid \theta_\mathcal{D}\right)\right\|^2, %+ 
     % \argmin_{\theta_\mathcal{E},\theta_\mathcal{D}}& \quad L \left( \bm{x}, \mathcal{D}\left( \mathcal{E}\left(\bm{x} \mid \theta_\mathcal{E}\right) \mid \theta_\mathcal{D}\right)\right)
     \\ \nonumber
     \mbox{s.t. } &\quad \forall \bm{x} \in \mathcal{N}:\mathcal{E}(\bm{x} | \theta_\mathcal{E}) \in B[0,R_1], \\ \nonumber
     & \quad \forall \bm{x} \in \mathcal{A}:\mathcal{E}(\bm{x} | \theta_\mathcal{E}) \notin B[0,R_2],
\end{align}

where $B[0,R_1]$ and $B[0,R_2]$ denote the closed balls, with radius $R_1$ and $R_2$, respectively, around the origin.
The learning objective is optimized using CGGD \citep{VanBaelen2023}.

To determine the CI for an input point $\bm{x}$ a similar procedure as for AE-DSVDD can be followed, with the difference being that for CGAE the origin is considered as the center instead of a point $\bm{c}$

\begin{equation}\label{eq:CGAE_CI}
    f_{CI}(\bm{x}) = \left\|\mathcal{E}(\bm{x}|\theta_\mathcal{E})-\bm{0}\right\|^2. 
\end{equation}

By using the constrained learning objective, it is expected that a (near) "optimal" threshold can be found somewhere in the interval $[R_1, R_2]$, as this lies between the normal and anomalous data.
Following this expectation, the threshold associated with CGAE is defined as
\begin{equation}\label{eq:threshold_CGAE}
    T:= R_1 + (R_2 - R_1)\frac{A}{N+A}.
\end{equation}
Do note that this threshold only depends on the amount of known normal and anomalous points that are available at training time and the choice of $R_1$ and $R_2$.
Furthermore, it is not influenced by the actual training of the model, as is the case for $T_{train}$ and $T_{sigmoid}$.
This threshold can then be used in \eqref{eq:AE-DSVDD_AD} to determine if a point should be considered normal or anomalous.

\section{Monotonically Constraint Guided Autoencoder}\label{sec:MCGAE}
In the application of condition monitoring of machines, it is natural to assume that the health of a machine cannot improve over time. Equivalently, the machine can over time only become less healthy or, equivalently, the CI can only increase over time. Therefore, it is desirable to predict a monotonically increasing sequence of norms for the encodings of the life time of a single machine, hereafter called a run. More specifically, for a given time series of data $\{\bm{x}_j\}_j$ (with the data chronologically ordered from left to right) consisting of normal and unlabeled data points of a single recording, it should hold that
    \begin{equation*}
        j_1 < j_2 \Rightarrow \left\|\mathcal{E}(\bm{x}_{j_1} \vert \theta_\mathcal{E})\right\| < \left\|\mathcal{E} (\bm{x}_{j_2} \vert \theta_\mathcal{E}) \right\|.
    \end{equation*}
To this end we propose Monotonically Constraint Guided AutoEncoder (MCGAE), which extends CGAE with a constraint that enforces this monotonous behavior.
This results in the extension of the optimization problem \eqref{eq:CGAE} to
\begin{align}\label{eq:MCGAE}
     \argmin_{\theta_\mathcal{E},\theta_\mathcal{D}}&\frac{1}{N}\sum\limits_{\bm{x} \in \mathcal{N}}\left\|\bm{x}-\mathcal{D}\left( \mathcal{E}\left(\bm{x} \mid \theta_\mathcal{E}\right) \mid \theta_\mathcal{D}\right)\right\|^2, %+ 
     \\ \label{eq:MCGAE:ConNormal} %\nonumber 
     \mbox{s.t. } &\forall \bm{x} \in \mathcal{N}:\mathcal{E}(\bm{x} | \theta_\mathcal{E}) \in B[0,R_1], \\  \label{eq:MCGAE:ConAnomalous}%\nonumber
     & \forall \bm{x} \in \mathcal{A}:\mathcal{E}(\bm{x} | \theta_\mathcal{E}) \notin B[0,R_2], \\  \label{eq:MCGAE:ConMonotonicity} %\nonumber
     & \forall r, \forall \bm{x}_{t_1}, \bm{x}_{t_2} \in \mathcal{S}^r: t_1 < t_2 < t^r_a \Rightarrow \left\|\mathcal{E}(\bm{x}_{t_1} \vert \theta_\mathcal{E})\right\| < \left\|\mathcal{E} (\bm{x}_{t_2} \vert \theta_\mathcal{E}) \right\|,
\end{align}
where $r$ denotes the considered run, $t^r_a$ is the time at which the first anomalous point occurs for run $r$, and $\mathcal{S}^r = \mathcal{N}^r \, \cup \, \mathcal{U}^r$, with $\mathcal{N}^r$ the set of normal points of run $r$, $\mathcal{U}^r=\{u^r_i\}_{i=1}^{U^r|}$ being the set of unlabeled points of run $r$, with $U^r$ the amount of unlabeled points of run $r$, and $\bm{x}_{t_i}$ is the sample on time $t_i$ of run $r$.
For simplicity, we omit the time index if it is not required.
Note that only points in a single run can be compared to each other for the monotonicity constraint, because, there is, in general, no relation between the encodings of certain time points from different runs. 
This optimization problem can be optimized using CGGD \citep{VanBaelen2023}. 
The latter framework enables training a neural network by minimizing a loss function while satisfying constraints.
This is realized by prioritizing satisfying the constraints over minimizing the objective function during every step of the optimization. 
In order to do so, it is required to  define the direction by which a certain model variable needs to be adjusted when it violates a corresponding constraint. 
This direction needs to be chosen such that when following it during the update the resulting updated model at least comes closer to the feasible region (that contains model solutions that satisfy the constraint). 
In case a constraint is satisfied the direction is considered to be $0$, otherwise, the different directions can be computed as follows.

Similar to CGAE, in case \eqref{eq:MCGAE:ConNormal} is not satisfied, the direction to update the model when normal training points are considered ($dir_\mathcal{N}$) is computed so that these points will be mapped inside of $B[0,R_1]$

\begin{equation*}
    dir_{\mathcal{N}}\left(\bm{x}, \mathcal{E}\right) := \frac{\mathcal{E}\left(\bm{x} \mid\theta_\mathcal{E}\right)}{\left\|\mathcal{E}\left(\bm{x} \mid\theta_\mathcal{E}\right)\right\|}.
\end{equation*}

Considering \eqref{eq:UpdateStepMCGAE} it can be observed that a scaled version of  direction ($dir_\mathcal{N}$) is added to the considered weight $\theta_j$ which will cause the solution to move towards the inside of the ball $B[0,R_1]$. 
The latter occurs as the computed direction will be followed in the opposite direction, due to the update procedure.
In this case, the computed direction is the vector from the origin to $\mathcal{E}\left(\bm{x} \mid\theta_\mathcal{E}\right)$, and hence, following this vector in the opposite direction causes the solution to move inwards.

In the previous equation a direction was calculated for a single point $\bm{x}$.  
To evaluate a set of points, this direction is aggregated, by summation.
As mentioned earlier, $dir_\mathcal{N}\left(\bm{x}, \mathcal{E}\right)$ is considered to be $0$ if the constraint is satisfied for point $\bm{x}$. 
%the direction for a point $\bm{x}$ is considered to be $0$ if the constraint is satisfied.
In case $\mathcal{E}\left(\bm{x} \mid\theta_\mathcal{E}\right)=0$, thus when the encoding of $\bm{x}$ is the origin, the direction to the origin is not computed, instead a random direction is chosen, that is, by sampling from a uniform distribution on the unit sphere.

In a similar manner, in case needed, to satisfy \eqref{eq:MCGAE:ConAnomalous}, the direction for the anomalous training points ($dir_\mathcal{A}$) is computed so that these points will be mapped outside of $B[0,R_2]$

\begin{equation*}
    dir_{\mathcal{A}}\left(\bm{x}, \mathcal{E}\right) := \frac{- \, \mathcal{E}\left(\bm{x} \mid\theta_\mathcal{E}\right)}{\left\|\mathcal{E}\left(\bm{x} \mid\theta_\mathcal{E}\right)\right\|}.
\end{equation*}

To satisfy \eqref{eq:MCGAE:ConMonotonicity}, the direction that causes the normal and unlabeled points to be mapped so that the norm of their corresponding encodings monotonically increases is computed as follows. 
For a given run, a function \textit{argsort($\cdot$)} is used that takes in a set of points and returns a vector of indices that are ranked in ascending order according to the norm of their corresponding encodings.

This vector of indices can then be compared to the expected order expressed by vector 
$\begin{bmatrix} 0 & ... &  |\mathcal{S}^r|-1 \end{bmatrix}$ by performing an element-wise subtraction to get the difference between the rankings. The resulting difference can then be normalized, using the L2 norm, to obtain the direction. More formally this becomes:

\begin{align*}
    dir_{mono}\left(\mathcal{S}^r, \mathcal{E}\right) := \frac{argsort\left(\left\{\left\|\mathcal{E}(\bm{x}|\theta_\mathcal{E})\right\|^2\mid \bm{x} \in \mathcal{S}^r\right\}\right)-\begin{bmatrix} 0 & ... &  |\mathcal{S}^r|-1 \end{bmatrix}}{\left\|argsort\left(\left\{\left\|\mathcal{E}(\bm{x}|\theta_\mathcal{E})\right\|^2\mid \bm{x} \in \mathcal{S}^r\right\}\right)-\begin{bmatrix} 0 & ... &  |\mathcal{S}^r|-1 \end{bmatrix}\right\|}.
\end{align*}

Note that, if the ranking for a point is far from the desired ranking, this point will more strongly influence the direction than if the ranking of a point is only 1 position different from the desired ranking.

The weights of a model can then be updated using the objective function and the directions for the different constraints

\begin{align} \label{eq:UpdateStepMCGAE}
    \theta_{j+1} = \theta_j - &\eta\left[\nabla L\left(\mathcal{N},\Phi(\mathcal{N})\right) \right. \\ \nonumber
    &\quad \quad+ R \, dir_\mathcal{N} \left(\mathcal{N}, \mathcal{E}\right) \max\left\{\left\|\nabla_e L\left(\mathcal{N},\Phi(\mathcal{N})\right)\right\|, \zeta\right\} \\ \nonumber
    &\quad \quad+ R \, dir_\mathcal{A} \left(\mathcal{A}, \mathcal{E}\right) \max\left\{\left\|\nabla_e L\left(\mathcal{N},\Phi(\mathcal{N})\right)\right\|, \zeta\right\} \\ \nonumber
    &\quad \quad\left.+ R \, dir_{mono} \left(\mathcal{S}^r, \mathcal{E}\right) \max\left\{\left\|\nabla_e L\left(\mathcal{N},\Phi(\mathcal{N})\right)\right\|, \zeta\right\} \right],
\end{align}

where $\Phi(\mathcal{N}) = \mathcal{D}\left( \mathcal{E}\left(\mathcal{N} \mid \theta_\mathcal{E}\right) \mid \theta_\mathcal{D}\right)$, $\eta$ is the learning rate, $R>1$ a fixed real number called the rescale factor, $\zeta>0$ a fixed real number, which is chosen small, $L$ is the objective function from \eqref{eq:MCGAE}, $\nabla_e L$ is the gradient of the loss function with respect to the encodings, and $\theta$ is the union of $\theta_\mathcal{E}$ and $\theta_\mathcal{D}$.
$\Phi(\mathcal{N})$ denotes the evaluation of the set of normal points $\mathcal{N}$ by the model.
This evaluation is done separately for each point in the set, and results in a new set that contains these evaluations.
The objective function from \eqref{eq:MCGAE} only uses normal data, this is denoted in \eqref{eq:UpdateStepMCGAE} by the use of the set of normal points $\mathcal{N}$.
As it is not feasible to use an entire run at once during training, this process will be performed on batches.
While this direction can be computed with only 2 points, it is expected that this will not perform well, and that more points are needed for this constraint to be used to compute a better direction.
Therefore, batches are used. 
Note that these are created in a stratified way, where it is guaranteed that, if a run is present in a batch, sufficient, e.g. 10 or more, points of that run are present in that batch.

As MCGAE is an extension of CGAE, the CI for an input point $\bm{x}$ can be determined using \eqref{eq:MCGAE}, and the associated threshold can also be acquired using \eqref{eq:threshold_CGAE}.
This threshold can then also be used in \eqref{eq:AE-DSVDD_AD} to distinguish between normal and anomalous points.

\section{Experimental setup}\label{sec:setup}
This section will discuss the datasets used in this work as well as the associated preprocessing, the model architectures and hyperparameter settings, and the relevant metrics.

\subsection{Datasets}\label{ssec:datasets}
The experiments in this work will be performed on two datasets. Firstly, a dataset containing vibration data from accelerated run-to-failure tests of bearings, hereafter called the Smart Maintenance (SM) dataset. Secondly, a dataset that contains acoustic data from accelerated run-to-failure tests of bearings, hereafter called Acoustic Bearing Monitoring (ABM) dataset.

\subsubsection{Smart Maintenance}\label{sssec:SM}
\begin{figure}[ht]
	\centering{\includegraphics[keepaspectratio=false, height=150px,width=0.75\columnwidth]{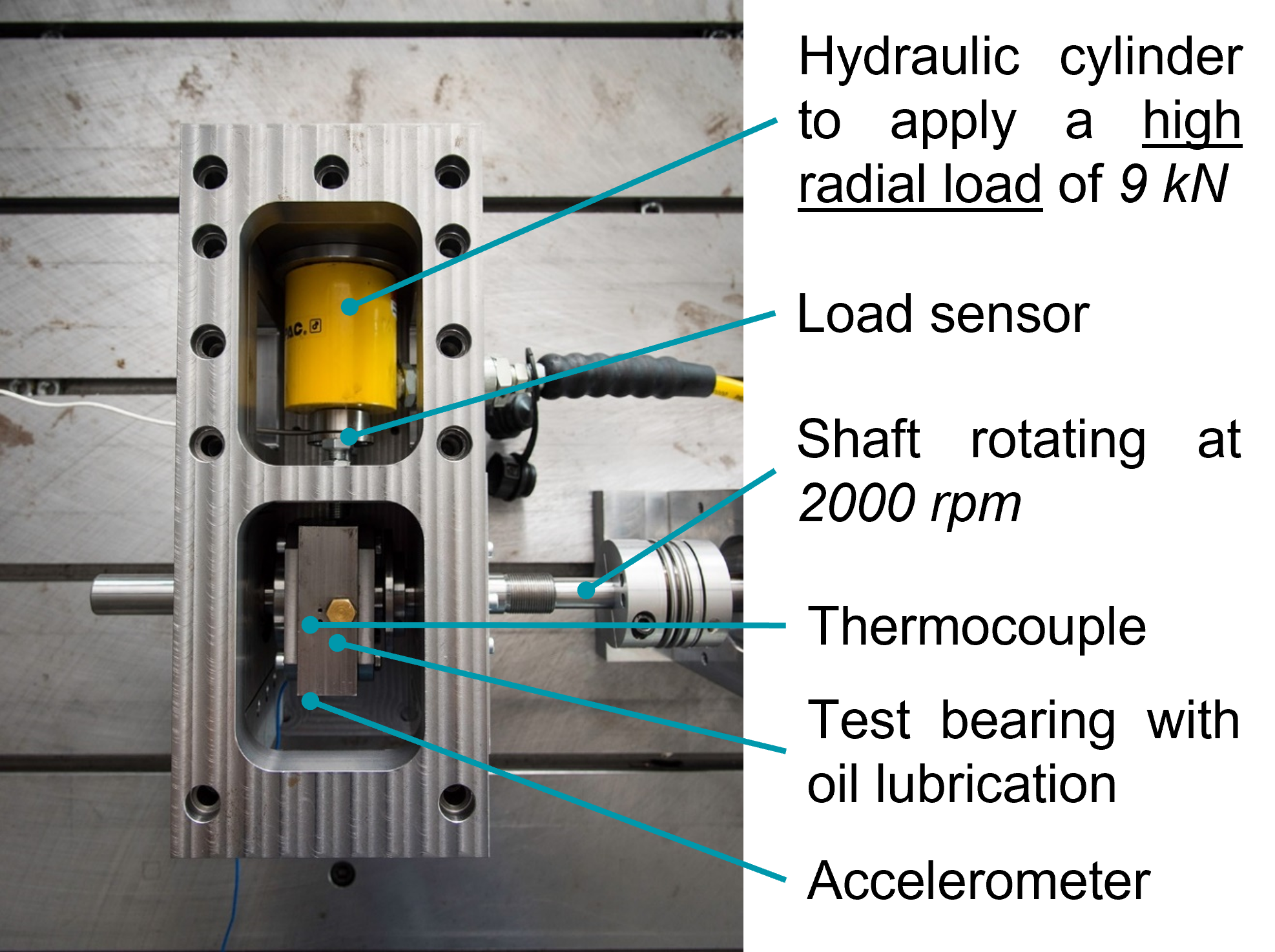}
	\caption{Example of a bearing test rig setup.}\label{fig:FM_Setup}}
\end{figure}

The Smart Maintenance (SM) dataset contains vibration data of a total of 70 accelerated run-to-failure tests, hereafter called runs, of bearings and was already used in \citep{Meire2023} and \citep{Meire2022}.
One of the 7 bearing rigs used to perform the tests is shown in Figure \ref{fig:FM_Setup}.

The runs were performed under fixed operation conditions, more specifically a high radial load of 9 kN and with the shaft rotating at 2000 rpm, and the stopping condition was set to 20g peak acceleration, which is the assumed end-of-life.
The high radial load and a very small indentation in the inner race of the bearing were used to significantly accelerate the bearing lifetime.
Radial accelerations were measured by an accelerometer, attached to the bearing housing, at a sampling frequency of 50 kHz.
As the aim is to work with a run-to-failure test under fixed conditions, data collected at a rotational speed lower than 2000 rpm during the initial run-up or during short measurement interruptions was omitted during the experiments.

Only data from 1 of the 7 bearing test rigs was selected to reduce the computing time needed for the ablation studies in this work. The data of this single setup contains 6 runs that start with a small initial indentation in the inner race of the bearing and end with a severe fatigue fault.
The length of these runs ranges from roughly 6600 to 14500 seconds, and the fault occurrence ranges from roughly 100 to 3000 seconds prior to the end of the run.

This dataset did not have an associated ground truth labelling, hence manual annotation was performed on each run, by inspecting the time-frequency representation of the data, to segment the data into three segments along the time axis.
This segmentation was done by determining cutoff points $p_h$ and $p_f$, with data prior to $p_h$ being considered as normal (although, at the start a small initial indentation was already present), data between $p_h$ and $p_f$ being considered as unlabeled, as the condition of the bearing is undefined, and data after $p_f$ being considered as anomalous.
The selection of these cutoff points is explained in detail in \citep{Meire2022}.

\subsubsection{Acoustic Bearing Monitoring}\label{sssec:ABM}
\begin{figure}[t]
    \centering
    \includegraphics[height=9cm]{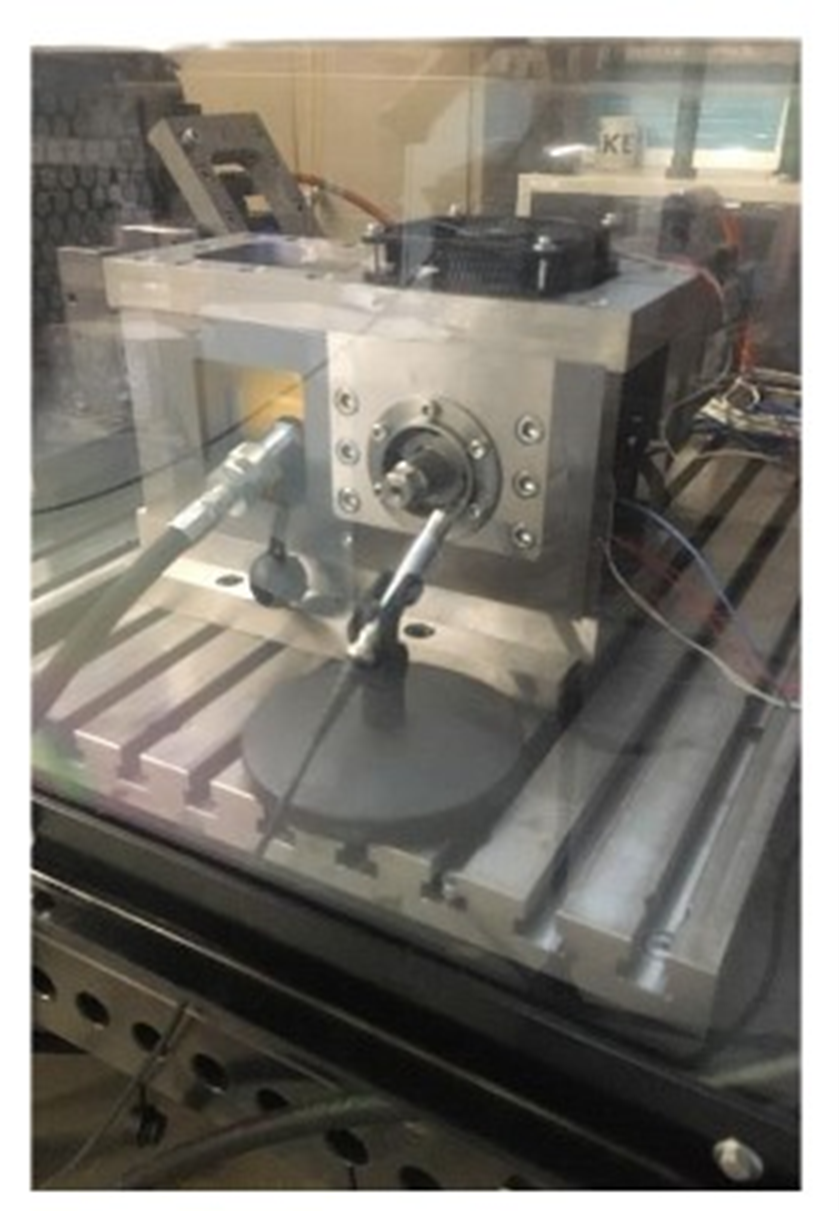}
  \caption{The microphone setup used in the bearing run-to-failure tests.}
  \label{fig:ACMON_setup}
\end{figure}

The Acoustic Bearing Monitoring (ABM) dataset contains both vibration and acoustic data of a total of 64 accelerated run-to-failure tests, hereafter called runs, of bearings and was collected on the same bearing test rig as shown in Figure \ref{fig:FM_Setup}.
The placement of the internal microphone can be seen in Figure \ref{fig:ACMON_setup}. 
This dataset was already used in \citep{Meire2023AL}.
In this work the data collected by the internal microphone, from runs with a fixed speed of 2000 rpm, and a fixed load were selected, resulting in 5 runs being retained. 
%In total 5 runs were retained. 
The length of these runs ranges from roughly 5000 to 18500 seconds, and the fault occurrence ranges from roughly 700 to 4300 seconds prior to the end of the run.

Data was captured with a sampling frequency of 50 kHz for each sensor.
Each run was accelerated in the same manner as the SM dataset, by creating a small indent on the inner race of the bearings in addition to a high radial load and the same stopping criteria of 20g peak accelerations as for the SM dataset was used.

Next to the raw data, a ground truth was also provided.
An important note is that this labelling was not based on the physical state of the bearings, but on analysis of the data collected by the internal microphone.
This dataset was also split into 3 segments along the time axis, similar to the SM dataset, by using two cutoff points $p_h$ and $p_f$.
$p_h$ was determined in the same way as for the SM dataset and $p_f$ was determined as the point in time where the faulty behavior was first detected.
Similar to the SM dataset, data prior to $p_h$ is considered as normal for the experiments, data between $p_h$ and $p_f$ as unlabeled, and data after $p_f$ is considered as anomalous.

\subsection{Preprocessing}\label{ssec:preprocessing}

The acceleration and acoustic signals are first transformed into a time-frequency representation, more specifically log mel spectra, as was done in \citep{Meire2022, Meire2023}.
This transformation is performed by first acquiring time-frequency spectra from the raw signals, using a window and hop size of 1 second.
From these spectra mel bands are then extracted, 64 and 512 bands for the ABM and SM datasets, respectively, and these are then log scaled.
Finally, 8 consecutive seconds were concatenated into an input frame, with shapes (64,8) and (512,8) for the ABM and SM dataset, respectively, which was then used as an input to the models.
{It was noticed that the distribution of the values of the mel spectra is different between runs.
Therefore, each run was separately standardized to have zero mean and unit variance prior to $p_h$.

\subsection{Experimental details}\label{ssec:details}

The models used in the experiments all use a similar AE architecture, consisting of an encoder and decoder.
The encoder uses convolutional blocks, consisting of a convolutional layer, followed by batch normalisation \citep{Batchnorm}, an activation function and finally a max pooling with stride and size of 2.
The decoder uses deconvolutional blocks, consisting of an upsampling with stride and size of 2, followed by a convolutional layer, batch normalisation and finally an activation function.
For the SM model the encoder consists of 3 convolutional blocks, with 64, 64, and 32 filters, respectively, and is followed by a fully connected (FC) layer with 8 neurons that serves as the encoding layer.
The decoder consists of a FC layer with 1024 neurons followed by 3 deconvolutional blocks, with 16, 32, and 32 filters respectively, and a final convolutional layer with a single filter is used as the output layer.
The model for the ABM datasets uses the same structure.
However, with an increased amount of filters and neurons.
The encoder uses 256, 256, and 128 filters in the convolutional blocks and 64 in the encoding FC layer, and the decoder uses 512 neurons in the first FC layer, followed by 128, 256, and 256 filters in the deconvolutional blocks, and the output layer remains the same.

The models for the SM model were trained for 300 epochs, and the models for the ABM dataset for 500 epochs, both with the Adam optimizer \citep{kingma2014} with a learning rate of $1e^{-3}$.
The best performing model was saved during training, and if the performance did not improve for 30 epochs, the learning rate was halved, up to a limit of $1e^{-6}$.
The training performance is determined by the learning objective associated with the algorithms, as described in Sections \ref{sec:methods} and \ref{sec:MCGAE}.
For CGAE and MCGAE the satisfaction ratio of the constraints, the amount of points for which the constraints are satisfied with respect to the total amount of points, is also considered.
An improvement in model performance for AE-DSVDD corresponds to the lowering of the learning objective on the validation set.
When considering CGAE or MCGAE, the improvement corresponds to both the lowering of the learning objective on the validation set and either an increase of the satisfaction ratio on the associated constraints or the satisfication ratio being above $95\%$. 

It was previously mentioned that the monotonicity constraint can only be applied to data from the same runs, and hence it is crucial that the runs in each batch are represented by multiple points.
To ensure this is the case, the batches were created in a structured manner.
This is done by first setting a number of runs that should be present in a batch and a number of points that should be selected from each run.
Based on these settings a batch is created by first randomly selecting runs and then selecting points from each of these runs.
For the latter, a stratified sampling was used that first selects a fixed number of points of the normal and, if applicable, anomalous data in a run and then another fixed number of points from the unlabeled data.
For the SM dataset each epoch a total of 80 batches were created in this way, with 5 runs in each batch, with 25 normal or anomalous points and an additional 10 unlabeled points for each of these runs.
For the ABM dataset the amount of batches was increased to 100. 
However, each batch only contained 4 runs, with the same amounts of points as the SM dataset.
Do note that this procedure will cause some points to be included in multiple batches in a single epoch.
However, this is not expected to have much impact on the training of the models.

\subsection{Metrics}\label{ssec:metrics}

The alternative methods studied in this paper will be compared in two ways: (i) the ability to discriminate between normal and anomalous behavior, and (ii) the quality of the CI in terms of its monotonicity.
The performance with regards to the discrimination between normal and anomalous data is evaluated using the balanced accuracy (BA) \eqref{eq:BA}, which is the mean of the recall obtained on each class.
While it is common to use the F1 score to evaluate discrimative performance, it was opted to use the BA instead in this work for two reasons.
Firstly, both the F1 score and BA were used in \citep{Meire2023}, where it could be seen that both showed similar results.
Secondly, the different runs show a varying amount of normal and anomalous data, which is not ideal for the F1 score as it is biased towards correctly classifying the positive class.
In this work anomalous data is considered as the positive class, leading to

\begin{align}
    \mbox{BA} &= \frac{1}{2}\left(\frac{TN}{(TN + FP)} + \frac{TP}{(TP + FN)}\right)\label{eq:BA}, 
\end{align}
where $TP$, $FP$, $TN$, and $FN$ correspond to the amount of true positives, false positives, true negatives, and false negatives, respectively. 
To evaluate the BA, a threshold is required.
Three thresholds, $T_{opt}, T_{train},$ and $T_{sigmoid}$ were described in Section \ref{ssec:ae-dsvdd} and will be used in the evaluation.

The second evaluation looks into the quality of the CI.
A high quality CI is expected to show an increasing trend up to a point where a fault occurs.
In \citep{Kim2016} and \citep{Meire2022} this was done by using the Spearman's $\rho$ \citep{Zwillinger1999}, which correlates the rank of two variables, A and B, and is calculated as follows,

\begin{equation}\label{eq:spearman}
\rho = \frac{cov(r(A), r(B))}{\sigma_{r(A)}\sigma_{r(B)}}
\end{equation}

with $r(A)$ and $r(B)$ being the ranking of $A$ and $B$, in ascending order, and $cov$ and $\sigma$ being the covariance and the standard deviation, respectively.
In the context of this work, $A$ and $B$ correspond to the CI and the time, respectively.

\section{Experiments and results}\label{sec:experimentsResults}
In this section a comparison is made between CGAE, AE-DSVDD, and our proposed method, MCGAE.
\subsection{Experiments}\label{ssec:experiments}
To compare alternatives, a leave-one-run-out scheme using both the SM and ABM dataset was employed.  
All experiments were repeated 3 times, using different seeds for model initialisation and batch generation.
As indicated earlier the selected subset of the SM dataset has 6 \textit{faulty} runs.
This results in 6 folds, as per the leave-one-run-out scheme, with each fold using 5 runs for training and validation, split $75\%$ and $25\%$, respectively, and the remaining run for testing.
The amount of anomalous data was incrementally increased by selecting the anomalous data from either 1, 2, 3, 4, or 5 runs used in the training and validation sets.
For each of these results the mean and standard deviation over the different folds was computed.

Methods were compared  both in terms of their ability to discriminate between normal and anomalous behavior and of the monotonicity (hence quality) of the generated CI. 
When evaluating the discriminative performance, a fixed threshold was chosen. 
For AE-DSVDD this was done by determining two thresholds based on training statistics, and for CGAE and MCGAE the threshold was determined using \eqref{eq:threshold_CGAE}, which also only uses information about the training data.
The quality of the CI will be evaluated in the same way for all models as will be explained in more detail later.

A similar setup was used for the ABM data set were 5 folds were used in the leave-one-run-out validation (because only 5 instead of 6 runs were available).

Next to the discussed experiments, an ablation study was performed to investigate the effect of applying the reconstruction AE loss in \eqref{eq:MCGAE} not only on normal data but also on unlabeled and anomalous data.
This was investigated as it is possible that reconstructing additional data could result in a better structured latent space.

\subsection{Results}\label{ssec:results}

This section will discuss the results of the performed experiments, as described in Section \ref{ssec:experiments}. 
Firstly, the discriminative performance with regards to the BA was evaluated.
Secondly, the quality of the generated CI was evaluated using the Spearman's $\rho$.
When evaluating the discriminative performance, a trivial predictor, that predicts all data as anomalous, is also provided for reference.
Next to the fixed thresholds and the trivial predictor, an "optimal" threshold is also considered.
Its corresponding discrimination performance can be considered as an upper bound.
As MCGAE extends CGAE, it is expected that the former outperforms the latter, mainly with regards to the quality of the CI, as this is the focus of the additional monotonicity constraint.

\subsubsection{Discriminative performance}

\begin{figure}[ht]
	\centering{\includegraphics[keepaspectratio=false,width=\columnwidth]{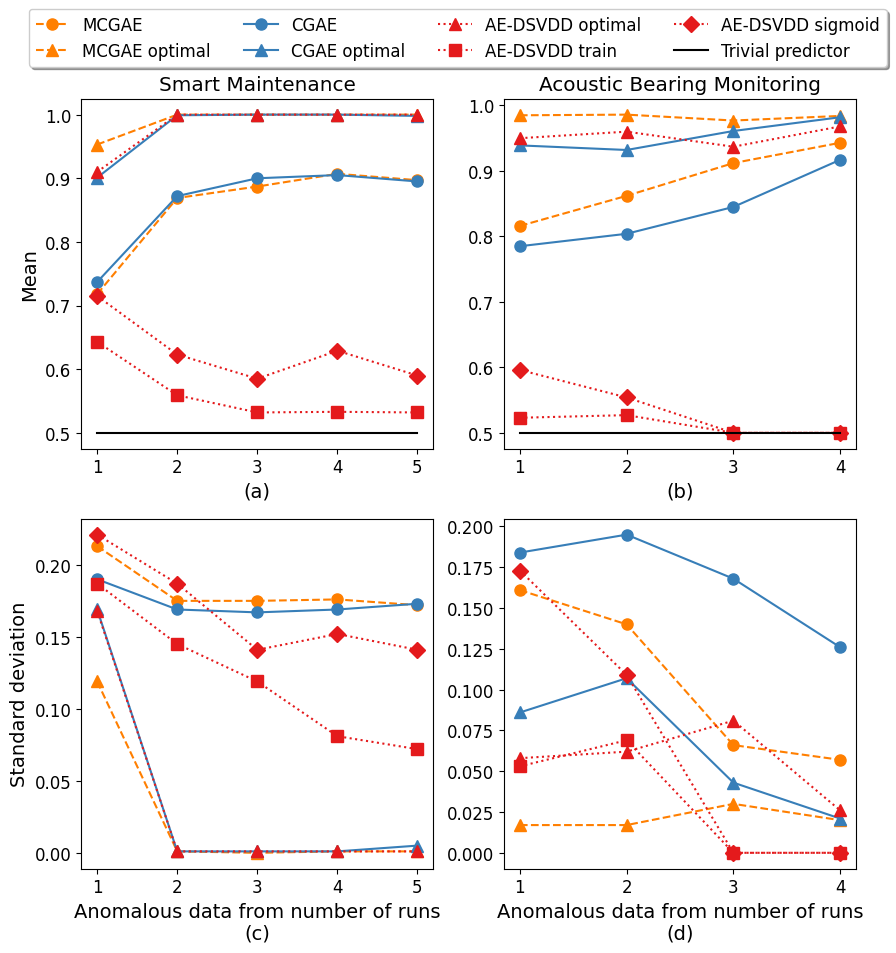}}
	\caption{The mean (a) and standard deviation (c) of the BA on the SM dataset, and the mean (b) and standard deviation (d) on the ABM dataset, for the different methods and different thresholds. The thresholds are $T_{train}$, $T_{sigmoid}$, and $T_{opt}$, as described in Section \ref{ssec:metrics}. The x-axis indicates from how many runs the anomalous data is used.}
	\label{fig:Method_comparison}
\end{figure}

The results of the comparison of the different methods in terms of the BA are shown in Figure \ref{fig:Method_comparison}.

When considering the results that are obtained when an optimal threshold was used it can be observed that MCGAE has equal or slightly improved performance compared to the alternatives. For the SM dataset this better performance is only noticeable when using anomalous data from 1 run, while for the ABM dataset MCGAE shows a consistent (near) perfect performance, and only when anomalous data from all 4 runs is used the performance of CGAE is similar.
CGAE does perform slightly worse than AE-DSVDD when anomalous data from 1 or 2 runs is used on the ABM dataset.
However, when more anomalous data is used it performs slightly better.
However, do remember that these thresholds are determined using the test data, and should be considered as the upper bound for the performance.

When not using an optimal threshold, both MCGAE and CGAE attain a performance that is significantly better than AE-DSVDD using a threshold based on training statistics, except for the case where the anomalous data from 1 run was used on the SM dataset.
Upon closer inspection of the generated CI, it was noticed that this lower score is attributed to 2 test runs where the CI for anomalous data is estimated to be lower than the threshold, whereas the other test runs perform closer to the behavior that is observed when more anomalous data is used.
This lower performance might be due to a combination of 2 reasons.
The first being that the model has only seen data from 1 type of anomaly, as only 1 run was used to select anomalous data.
The second being that it is possible that the model was trained with a limited amount of anomalous data for 1 or more folds, as anomalous data is selected from only 1 run and some runs only contain a limited amount of anomalous data.
When using anomalous data from more runs this is no longer the case, and the BA remains relatively constant.
This phenomenon was less prevalent on the ABM dataset, although a clear positive trend can be seen when using anomalous data from more runs, both for the SM and ABM dataset.
This trend is mainly attributed to 2 test runs where the estimated CI exceeds the chosen threshold closer to $p_f$ when using more anomalous data.

When evaluating the performance of AE-DSVDD when using a threshold based on training statistics, it can be seen that the performance is generally lower than CGAE and MCGAE, except for when anomalous data from only 1 run is used on the SM dataset.
Upon closer investigation, it was noticed that, mostly, only a few test runs, 2 for the SM dataset and 1 for the ABM dataset, did not show the same performance as the trivial predictor, where all predictions are considered anomalous, and the change in the metric is thus attributed to these runs.

The standard deviations show similar trends between the methods.
It can be seen that, in general, the standard deviations are lower for AE-DSVDD when using training statistics than for CGAE and MCGAE.
However, this is due to most of the test runs performing very similar to the trivial predictor.

\subsubsection{Condition indicator}

Next to discrimination performance also the quality of the CI estimation is assessed. More specifically, the quality of the CI is evaluated using the Spearman's $\rho$, described in Section \ref{ssec:metrics}, for which the results can be found in Table \ref{tab:CI_quality_test}.

\begin{table}[ht]
\caption{The Spearman's $\rho$ results for the both datasets on the test set. The best results for each amount of anomalous data are shown in bold.}
\label{tab:CI_quality_test}
    \begin{tabular}{cccc}
    \toprule
    \multicolumn{4}{c}{SM data set} \\ \cmidrule{1-4} 
    Anomalous data from number of runs &  MCGAE & CGAE  & AE-DSVDD  \\  \cmidrule{1-4} 
    \multicolumn{1}{l|}{1}      & \textbf{0.658 ± 0.304}     &     0.439 ± 0.229  & 0.284 ± 0.285  \\
    \multicolumn{1}{l|}{2}      & \textbf{0.667 ± 0.233}     &     0.439 ± 0.419    & 0.205 ± 0.597  \\
    \multicolumn{1}{l|}{3}      & \textbf{0.588 ± 0.369}     &     0.538 ± 0.432 & 0.238 ± 0.415  \\
    \multicolumn{1}{l|}{4}      & \textbf{0.582 ± 0.377}      &    0.515 ± 0.452     &  0.475 ± 0.319  \\
    \multicolumn{1}{l|}{5}     & \textbf{0.558 ± 0.403}      &    0.475 ± 0.467   & 0.518 ± 0.350  \\ 
    \cmidrule{1-4} 
    \multicolumn{4}{c}{ABM data set} \\ \cmidrule{1-4} 
    Anomalous data from number of runs &  MCGAE & CGAE          & AE-DSVDD  \\ \cmidrule{1-4}  
    \multicolumn{1}{l|}{1}      & \textbf{0.228 ± 0.308 }        & 0.157 ± 0.247     & 0.184 ± 0.456  \\
     \multicolumn{1}{l|}{2}      & \textbf{0.249 ± 0.285}         & 0.099 ± 0.251   & 0.116 ± 0.428  \\
     \multicolumn{1}{l|}{3}      &  \textbf{0.213 ± 0.451}         & 0.034 ± 0.255     & -0.052 ± 0.516  \\
     \multicolumn{1}{l|}{4}    & \textbf{0.174 ± 0.438}         & 0.007 ± 0.319     & 0.053 ± 0.400 \\
    \bottomrule
    \end{tabular}
\end{table}

It can be seen that MCGAE shows the highest Spearman's $\rho$ overall for both datasets.
When looking at the results for MCGAE for the SM dataset more closely, a better performance is noticed when anomalous data from only 1 or 2 is used.
The lower performance when using more anomalous data is caused by 2 test runs performing poorer while the other test runs show a more constant performance.
This slight decrease in the Spearman's $\rho$ when anomalous data from more runs is used can also be seen on the ABM dataset, albeit to a lesser extent.

While the behavior of CGAE and AE-DSVDD is similar, it can be seen that the former does perform slightly better when considering the SM dataset.
However, as was also the case for MCGAE, the performance difference is not always clear when considering the ABM dataset.
This different behavior between the datasets could be due to the ABM dataset being more noisy, as it contains acoustic data, in comparison to the SM dataset, which contains vibration data.

While MCGAE does show a better performance, it also shows a decreasing trend when using anomalous data from more runs.
As mentioned earlier, this is mainly due to 2 test runs showing a strong decrease in Spearman's $\rho$.
However, when observing the other methods for the SM dataset, an increasing trend can be seen.
This could be due to the nature of the constraint not being simple to generalize to different unseen test runs, as an increasing trend in a CI over time is strong tied to the run itself, and bearings can show different degradation patterns, even when operating under the same conditions \citep{Sikorska2011}.
The overall lower Spearman's $\rho$ on the ABM dataset is likely also due to poor generalisation, which is further worsened due to increased noise in the dataset.

To investigate the effect of the added constraint without the difficulties of the generalisation, the Spearman's $\rho$ is also evaluated on the training set, and the corresponding results are shown in Table \ref{tab:CI_quality_train}.

\begin{table}[ht]
\caption{The Spearman's $\rho$ results for the both datasets on the training set. The best results for each amount of anomalous data are shown in bold.}
\label{tab:CI_quality_train}
    \begin{tabular}{cccc}
    \toprule
    \multicolumn{4}{c}{Smart Maintenance} \\ \cmidrule{1-4} 
    Anomalous data from number of runs &  MCGAE & CGAE  & AE-DSVDD  \\  \cmidrule{1-4} 
    \multicolumn{1}{l|}{1}      & \textbf{0.891 ± 0.062}         & 0.235 ± 0.554    & 0.396 ± 0.437  \\
    \multicolumn{1}{l|}{2}      & \textbf{0.926 ± 0.040}         & 0.536 ± 0.445    & 0.547 ± 0.305  \\
    \multicolumn{1}{l|}{3}      & \textbf{0.931 ± 0.037}         & 0.626 ± 0.336    & 0.579 ± 0.343  \\
    \multicolumn{1}{l|}{4}      & \textbf{0.943 ± 0.029}         & 0.677 ± 0.341    & 0.608 ± 0.254  \\
    \multicolumn{1}{l|}{5}    & \textbf{0.945 ± 0.030}         & 0.631 ± 0.361    & 0.658 ± 0.211  \\  
    \cmidrule{1-4} 
    \multicolumn{4}{c}{Acoustic Bearing Monitoring} \\ \cmidrule{1-4} 
    Anomalous data from number of runs &  MCGAE & CGAE          & AE-DSVDD  \\ \cmidrule{1-4}  
    \multicolumn{1}{l|}{1}      & \textbf{0.860 ± 0.089}         & 0.216 ± 0.341 & 0.463 ± 0.318  \\
    \multicolumn{1}{l|}{2}      & \textbf{0.872 ± 0.091}         & 0.266 ± 0.354     & 0.473 ± 0.301  \\
    \multicolumn{1}{l|}{3}      & \textbf{0.883 ± 0.096}         & 0.355 ± 0.274     & 0.553 ± 0.253  \\
    \multicolumn{1}{l|}{4}    & \textbf{0.898 ± 0.080}         & 0.433 ± 0.221     & 0.555 ± 0.251  \\ 
    \bottomrule
    \end{tabular}
\end{table}

It is clear that MCGAE attains the highest Spearman's $\rho$ when considering the runs used during training, and that the difference with the other methods is significantly larger than on the test set.
This indicates the effectiveness of the constraint and that the generalization to an unseen test run is indeed not simple.
When comparing the difference in performance for MCGAE on the training and test set between the SM and ABM dataset, it is clear that this difference is noticeably larger for the latter.
This indicates that the generalization is indeed more difficult on the ABM dataset, likely due to the increased acoustic noise in the dataset, as mentioned earlier.

Other interesting observations, on both datasets, are the performance when anomalous data from only 1 run is used, as the Spearman's $\rho$ is already good, and the increasing performance when anomalous data from more runs is used.
The former is likely due to the influence of the constraint and the latter due to a combination of the constraint and the increased amount of anomalous data providing a better structure in the latent space.
The increasing performance is also noticed for CGAE and AE-DSVDD, although their performance remains noticeably lower than that of MCGAE.
It is also interesting that the Spearman's $\rho$ shows a somewhat stronger increasing trend for CGAE than for AE-DSVDD.

In \citep{Meire2023}, it was evaluated how similar the fixed thresholds are to an "optimal" threshold.
As the performance of the latter can be considered an upper bound, a more similar threshold can also be considered a better threshold.
While this is not the main focus of this work, the same evaluation was performed, and the results were similar as those shown in \citep{Meire2023}.
The detailed results are shown and discussed in Appendix \ref{appendix:ths_methods}.

\section{Ablation study}\label{sec:ablation}

This section discusses the results obtained when reconstructing additional data next to only the normal data.
Experiments were performed with 4 different combinations of reconstructed data: (i) only the normal data (\textit{MCGAE-n}), these are the same as in Section \ref{ssec:results}, (ii) additionally reconstructing unlabeled data (\textit{MCGAE-nu}), (iii) additionally reconstructing anomalous data (\textit{MCGAE-na}), and (iv) additionally reconstructing both unlabeled and anomalous data (\textit{MCGAE-nua}).

\subsection{Discriminative performance}

\begin{figure}[ht]
	\centering{\includegraphics[keepaspectratio=false,width=\columnwidth]{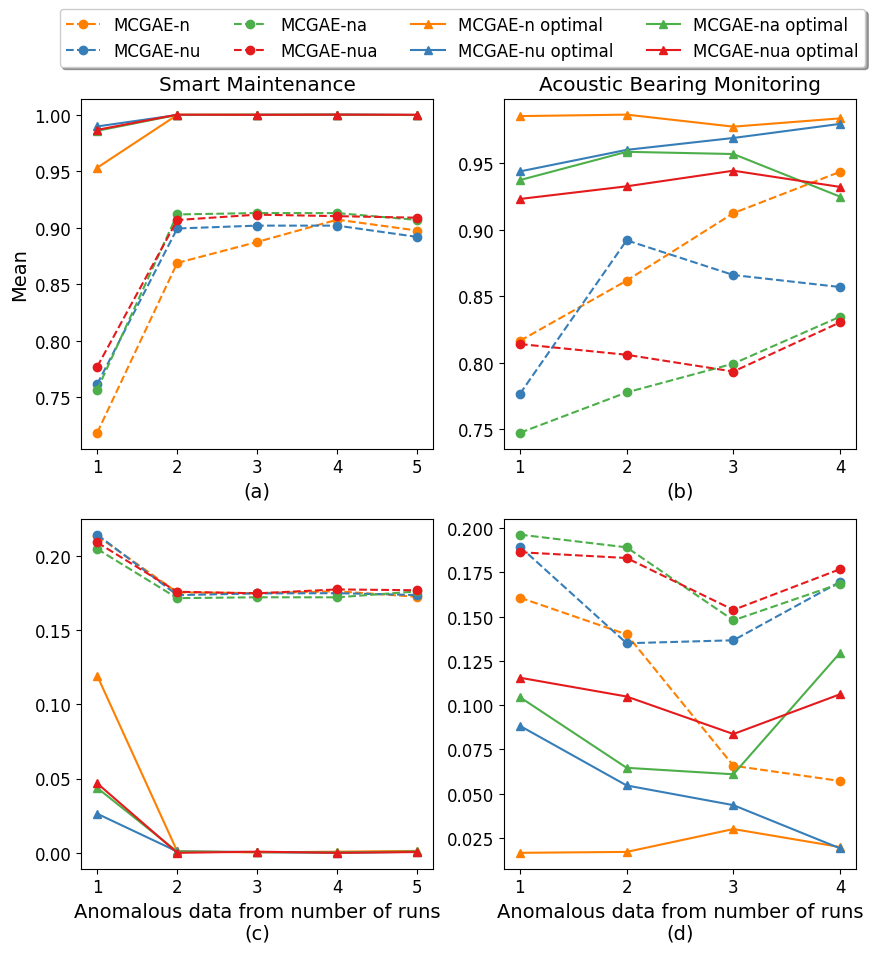}}
	\caption{The mean (a) and standard deviation (c) of the BA on the SM dataset, and the mean (b) and standard deviation (d) on the ABM dataset, for the different reconstructions. The x-axis indicates the number runs from which anomalous data is used.}
	\label{fig:Ablation_comparison}
\end{figure}

The results of the comparison of the different reconstruction based on the BA is shown in Figure \ref{fig:Ablation_comparison}.
It can be seen that the results on the SM and ABM datasets are different.
When considering a chosen threshold on the SM dataset, it is clear that there are no large differences in performance for the different reconstructions.
However, MCGAE-n seems to perform slightly worse, mainly when using anomalous data from up to 3 runs.
This is due to 1 or 2 test runs, across the different folds and seeds, performing worse, while the other test runs perform the same.
It also seems that MCGAE-na and MCGAE-nua perform slightly better overall.
However, the difference is very minimal.
When considering an optimal threshold this performance difference is even further reduced, as all reconstructions attain a near perfect BA, except for when anomalous data from 1 run is used.
This indicates that the lower performance of MCGAE-n using the chosen threshold is due to a mismatch between the chosen and optimal threshold, and not due to the model performing worse.

When considering a chosen threshold on the ABM dataset, it is indicated that MCGAE-n and MCGAE-nu perform better overall, except when anomalous data from 1 or 2 runs is used.
This behavior is slightly different when an optimal threshold is considered, as MCGAE-n outperforms MCGAE-nu when anomalous data from less than 3 runs is used.
It is indicated that MCGAE-na performs the worst, when considering a chosen threshold, although the difference with MCGAE-nua is minimal when anomalous data from more than 2 runs is used.
When selecting an optimal threshold both MCGAE-na and MCGAE-nua also perform worse than MCGAE-n and MCGAE-nu.
This could be due to the CI estimated by either MCGAE-na or MCGAE-nua, and to a lesser extent MCGAE-nu, being less distinct for the normal and anomalous data in comparison to MCGAE-n.
The cause of this lesser distinction is mainly split into 2 parts.
Firstly, it seems that reconstructing unlabeled data, MCGAE-nu and MCGAE-nua, more strongly influences the normal data, and causes the estimated CI of this data to increase, possibly due to the monotonicity constraint enforcing the CI of unlabeled data to be higher, while the additional reconstruction likely influences the normal data as well, as these are mostly similar.
Secondly, reconstructing the anomalous data, MCGAE-na and MCGAE-nua, shows a small influence on the normal data, possibly due to a few similarities in the data close to $p_f$, and a strong effect on the estimated CI of the anomalous data, causing this CI to be noticeably lower.

\subsection{Condition indicator}

\begin{table}[ht]
\caption{The Spearman's $\rho$ results for the both datasets on the test set. The best results for each amount of anomalous data are shown in bold.}
\label{tab:Ablation_CI_quality_test}
    \begin{tabular}{ccccc}
    \toprule
    \multicolumn{5}{c}{Smart Maintenance} \\ \cmidrule{1-5} 
     \makecell{Anomalous data from \\ number of runs} &  MCGAE-n & MCGAE-nu  & MCGAE-na           & MCGAE-nua   \\ \cmidrule{1-5} 
    \multicolumn{1}{l|}{1}      & \textbf{0.658 ± 0.304}         & 0.601 ± 0.343    & 0.516 ± 0.401 & 0.569 ± 0.364  \\
    \multicolumn{1}{l|}{2}      & \textbf{0.677 ± 0.233}         & 0.622 ± 0.283    & 0.463 ± 0.035 & 0.479 ± 0.347  \\
    \multicolumn{1}{l|}{3}      & 0.588 ± 0.369         & \textbf{0.631 ± 0.330}    & 0.513 ± 0.377 & 0.476 ± 0.347  \\
    \multicolumn{1}{l|}{4}    & 0.582 ± 0.377         & \textbf{0.620 ± 0.315}    & 0.497 ± 0.389 & 0.539 ± 0.346  \\
    \multicolumn{1}{l|}{5}    & \textbf{0.558 ± 0.403}         & \textbf{0.553 ± 0.357}    & \textbf{0.540 ± 0.343} & \textbf{0.569 ± 0.306}  \\       \cmidrule{1-5} 
    \multicolumn{4}{c}{Acoustic Bearing Monitoring} \\ \cmidrule{1-5}
    \makecell{Anomalous data from \\ number of runs} &  MCGAE-n & MCGAE-nu  & MCGAE-na           & MCGAE-nua   \\ \cmidrule{1-5} 
    \multicolumn{1}{l|}{1}      & \textbf{0.228 ± 0.308}         & 0.160 ± 0.325    & 0.175 ± 0.280 & 0.066 ± 0.254  \\
    \multicolumn{1}{l|}{2}      & \textbf{0.249 ± 0.285}         & 0.162 ± 0.315    & 0.153 ± 0.255 & 0.112 ± 0.332  \\
    \multicolumn{1}{l|}{3}      & \textbf{0.213 ± 0.451}         & 0.175 ± 0.280    & 0.100 ± 0.366 & 0.086 ± 0.352  \\
    \multicolumn{1}{l|}{4}    & 0.174 ± 0.438         & \textbf{0.212 ± 0.408}    & 0.162 ± 0.320 & 0.151 ± 0.382  \\
    \bottomrule
    \end{tabular}
\end{table}

The evaluation of the quality of the CI using the Spearman's $\rho$ for the different reconstructions can be found in Table \ref{tab:Ablation_CI_quality_test}.
Overall, it can be seen that MCGAE-n and MCGAE-nu show a higher Spearman's $\rho$ in comparison to MCGAE-na and MCGAE-nu, except for the SM dataset when using anomalous data from all runs.
As mentioned in Section \ref{ssec:results} the decreasing trend for MCGAE-n on the SM dataset is mostly due to 2 test runs.
For MCGAE-nu the Spearman's $\rho$ shows a more increasing trend, expect for the SM dataset when anomalous data from all runs is used, where a noticeable decrease in performance can be seen.
This is mainly due to a significantly lower performance of 4 test runs, across all folds and seeds, while the rest of the runs show a more stagnant Spearman's $\rho$.
A small difference between the results for both datasets can be seen for MCGAE-na and MCGAE-nua.
When considering the SM dataset, both reconstructions perform quite similarly.
However, on the ABM dataset MCGAE-nua performs worse than MCGAE-na.
As was also done when comparing the different methods, the quality of the CI on the training set will also be evaluated to omit the potentially difficult generalisation.
The results are shown in Table \ref{tab:Ablation_CI_quality_train}.

\begin{table}[ht]
\caption{The Spearman's $\rho$ results for the both datasets on the training set. The best results for each amount of anomalous data are shown in bold.}
\label{tab:Ablation_CI_quality_train}
    \begin{tabular}{ccccc}
    \toprule
    \multicolumn{5}{c}{Smart Maintenance} \\ \cmidrule{1-5} 
     \makecell{Anomalous data from \\ number of runs} &  MCGAE-n & MCGAE-nu  & MCGAE-na           & MCGAE-nua   \\ \cmidrule{1-5} 
    \multicolumn{1}{l|}{1}      & 0.891 ± 0.062         & \textbf{0.906 ± 0.086}    & 0.870 ± 0.088 & 0.877 ± 0.100  \\
    \multicolumn{1}{l|}{2}      & 0.926 ± 0.040         & \textbf{0.939 ± 0.044}    & 0.899 ± 0.066 & 0.910 ± 0.082  \\
    \multicolumn{1}{l|}{3}      & 0.931 ± 0.037         & \textbf{0.952 ± 0.026}    & 0.927 ± 0.049 & 0.931 ± 0.052  \\
    \multicolumn{1}{l|}{4}    & 0.943 ± 0.029         & \textbf{0.955 ± 0.035}    & 0.944 ± 0.036 & 0.922 ± 0.129  \\
    \multicolumn{1}{l|}{5}    & 0.945 ± 0.030         & \textbf{0.965 ± 0.018}    & 0.954 ± 0.024 & 0.956 ± 0.032  \\       \cmidrule{1-5} 
    \multicolumn{4}{c}{Acoustic Bearing Monitoring} \\ \cmidrule{1-5}
    \makecell{Anomalous data from \\ number of runs} &  MCGAE-n & MCGAE-nu  & MCGAE-na           & MCGAE-nua   \\ \cmidrule{1-5} 
    \multicolumn{1}{l|}{1}      & 0.860 ± 0.089         & 0.938 ± 0.052 & 0.886 ± 0.082 & \textbf{0.951 ± 0.037}  \\
    \multicolumn{1}{l|}{2}      & 0.872 ± 0.091         & \textbf{0.946 ± 0.052}    & 0.907 ± 0.066 & \textbf{0.951 ± 0.050}  \\
    \multicolumn{1}{l|}{3}      & 0.883 ± 0.096         & \textbf{0.963 ± 0.039}    & 0.925 ± 0.061 & 0.955 ± 0.049  \\
    \multicolumn{1}{l|}{4}    & 0.898 ± 0.080         & \textbf{0.963 ± 0.035}    & 0.936 ± 0.053 & \textbf{0.968 ± 0.024}  \\
    \bottomrule
    \end{tabular}
\end{table}

It can be seen that these results show a different behavior than those on the test set.
For the SM dataset, MCGAE-nu shows the overall highest Spearman's $\rho$ instead of MCGAE-n.
Both MCGAE-na and MCGAE-nua still show a slighly lower Spearman's $\rho$, except for when anomalous data from all runs is used, although the difference is not always significant.
For the ABM dataset, MCGAE-n shows the lowest Spearman's $\rho$ out of the different reconstructions, and MCGAE-nu and MCGAE-nua show the highest Spearman's $\rho$.
This indicates that additionally reconstructing the unlabeled data seems to improve the effect on the constraint on the training set, likely due to a similarity between the normal data and most of the unlabeled data.
The influence of reconstructing anomalous data is not as clear, for the SM dataset there is no real benefit, while for the ABM dataset there is some gain in Spearman's $\rho$ in comparison to MCGAE-n.
However, MCGAE-nua performs similar to MCGAE-nu, indicating that the reconstruction of the unlabeled dataset has a stronger influence on the constraint.

As was done in Section \ref{ssec:results}, the evaluation with regards to how similar the fixed threshold are to an "optimal" threshold was also performed for this ablation study, and the detailed results are shown and discussed in Appendix \ref{appendix:ths_reconstructions}.

\section{Conclusion and future work}\label{sec:conclusion}

In this work an extension to the CGAE algorithm, MCGAE, was proposed to jointly optimize a model for CI estimation and AD.
This extension adds a constraint that enforces a monotonic behavior in the estimated CI, while retaining the ability of CGAE to discriminate between normal and abnormal data.
The proposed MCGAE algorithm was evaluated and compared to AE-DSVDD and CGAE in terms of the discriminative performance and the quality of the CI.
This comparison was performed on two datasets, the SM dataset containing vibration data from run-to-failure tests of bearings, and the ABM dataset containing acoustic data from similar run-to-failure tests.
Next to the comparison with AE-DSVDD and CGAE, an ablation study was also performed to evaluate the reconstruction of unlabeled and/or anomalous data.

The results show that MCGAE attains a similar or slightly better discriminative performance than CGAE, depending on the considered dataset, while also providing a higher quality CI.
However, it should be noted that, while increasing the amount of anomalous data improves the discriminative performance, it does result in a decrease in the quality of the CI.
It is possible that this is due to the monotonicity constraint not being simple to generalize to unseen runs, as the increasing trend in CI over the life of a bearing is strongly tied to the bearing itself.
Additionally, as the runs are highly accelerated, the variation in the degradation trends is also higher.
Specifically for the ABM dataset, as acoustic data is more prone to noise than vibration data, this could result in the lower monotonicity.
When evaluating the quality of the CI on the training set, to investigate the effect of the monotonicity constraint without the generalization, MCGAE showed a noticeably higher quality of the CI, and increasing the amount of anomalous data also improved the CI.

The results of the ablation study show that, depending on the dataset, the discriminative performance is affected differently based on what data is reconstructed.
On the SM dataset, reconstructing additional data shows a slight improvement, while on the ABM dataset it noticeably lowers the performance.
The quality of the CI on an unseen run is highest when reconstructing either only normal data, or normal and unlabeled data.
However, when evaluating the quality of the CI on the training set, additionally reconstructing unlabeled data does show a clear increase in performance.

A possibility for future research is the improvement of the function \eqref{eq:threshold_CGAE} that determines the threshold for (M)CGAE, either through including some (limited) test data for calibration, a dynamic thresholding mechanism, a more advanced function, or incorporating insights from e.g. industry, such as considering false negatives to be more important than false positives, or vice versa.

Another possibility could be investigating whether a MCGAE model could be fine-tuned to an unseen test set by using some data from this set.
This could be seen as a more advanced calibration in comparison to what was mentioned in the previous point.

\appendix

\section{Relative difference in thresholds for the different methods}\label{appendix:ths_methods}

This appendix will discuss the (relative) difference between an "optimal" threshold and other thresholding methods for the comparison between AE-DSVDD, CGAE, and MCGAE.

\subsection{Metric}
To perform AD based on an estimated CI a threshold is need.
In Section \ref{ssec:ae-dsvdd} 2 thresholds based on training data/statistics, $T_{train}$ and $T_{sigmoid}$, and 1 threshold based on testing, $T_{opt}$, were described.
As $T_{opt}$ uses the testing data, it serves as an upper bound, and it is expected that other thresholds that are similar to it should also perform well.
Hence the relative difference between $T_{opt}$ and the thresholds that are determined using the training data is evaluated.
This relative difference is defined by the function described in \citep{Meire2023}

\begin{equation}\label{eq:difference}
    T_{diff,m}:\mathbb{R}^{+} \to \mathbb{R}: T \mapsto \frac{T_{opt} - T}{T_{opt}}, 
\end{equation}
where $m$ is the name of the method that is under evaluation, and $\mathbb{R}^+$ the set of positive real numbers.
Do note that this difference can be a negative value, which indicates that a chosen threshold is higher than the optimal threshold.

\subsection{Results}

\begin{figure}[ht]	\centering{\includegraphics[keepaspectratio=false,width=\columnwidth]{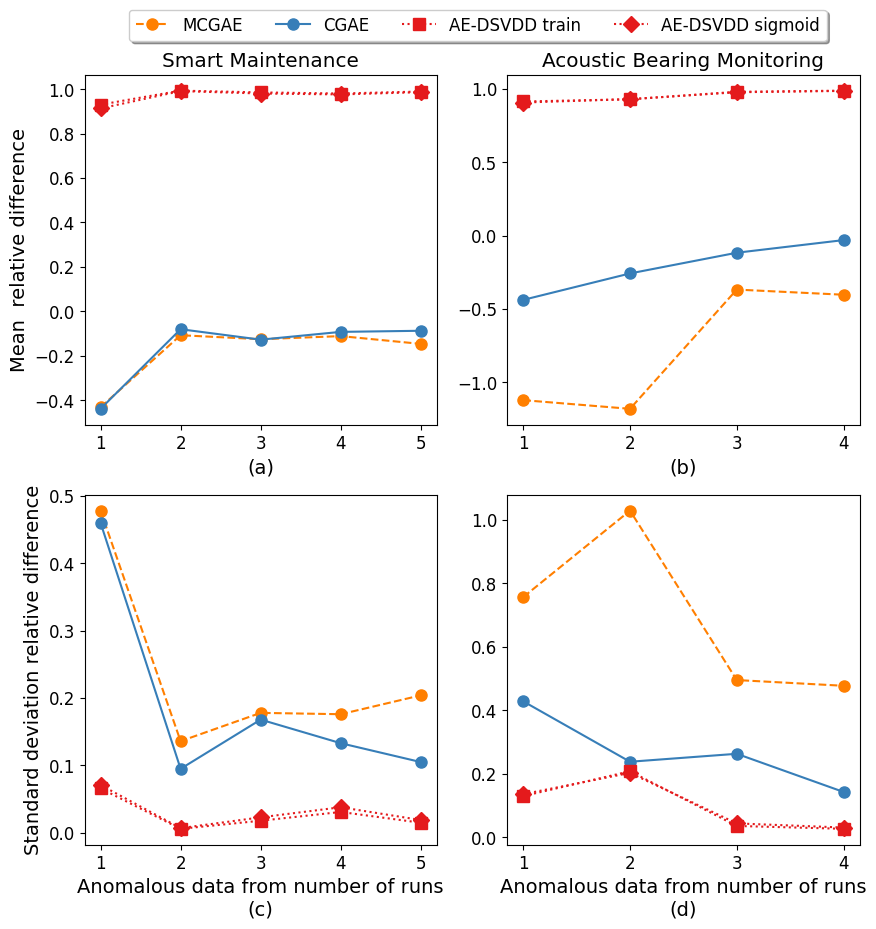}}
	\caption{The mean (a) and standard deviation (c) of the relative difference in norm between the different thresholds and the corresponding optimal threshold for the BA on the SM dataset, as well as the ABM dataset (b) and (d), for the different methods. The thresholds are $T_{train}$, $T_{sigmoid}$, as described in Section \ref{ssec:ae-dsvdd}.}
	\label{fig:thresh_comparison}
\end{figure}

The results for the comparison between the thresholds based on training statistics and an optimal threshold for the different methods are shown in Figure \ref{fig:thresh_comparison}.
The SM and ABM datasets show some similarities and some differences with regards to these results.
For both datasets AE-DSVDD shows a relative difference that is close to 1 and a low value for the standard deviation.
This means that the optimal threshold is (significantly) higher than the thresholds based on the training statistics.
This is not unexpected as the normal data is often mapped quite close to the center $c$ during training, causing a relatively low threshold when using statistics from this training, whereas the estimated CI on an unseen run tends to be higher, causing the optimal threshold to also be higher.
An example of this difference between the estimated CI for normal data from the training and test set can be seen in Figure \ref{fig:Example_CI_hist_FM}.

\begin{figure}[ht]
	\centering{\includegraphics[keepaspectratio=false,width=\columnwidth]{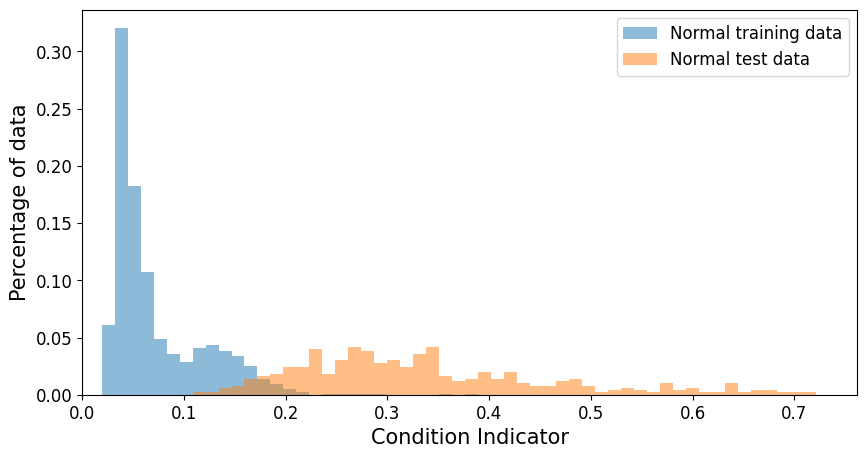}}
	\caption{A histogram of the CI estimated by AE-DSVDD for the normal data in the training and test set when using anomalous data from 1 run. The data from all folds was combined to create the histogram and the x-axis was stopped at twice the maximum CI of the normal training data for better visualisation.}
	\label{fig:Example_CI_hist_FM}
\end{figure}

For both datasets MCGAE and CGAE show a negative relative difference, which means that the optimal threshold is lower than the fixed threshold.
Both methods also show a lower absolute value, except for the ABM dataset with anomalous data from 1 or 2 runs, indicating that the chosen thresholds are, overall, closer to the optimal threshold.
For the SM dataset, it can be seen that the chosen threshold becomes closer to the optimal threshold when anomalous data from more than 1 run is used, for both CGAE and MCGAE.
This behavior could also be seen with regards to the discriminative performance.
The results on the ABM datasets are different, as both CGAE and MCGAE show a large relative difference between the optimal and chosen threshold.
It can also be seen that the mean relative difference for MCGAE is noticeably lower than that of CGAE, especially when anomalous data from 1 or 2 runs is used..
However, it should be noted that both methods show that the relative difference decreases when anomalous data from more runs is used.
This could also explain the increasing discriminative performance.
It might seem that CGAE shows a lower discriminative performance while the chosen threshold is closer to the optimal threshold than MCGAE, for the ABM dataset, it should be taken into account that the optimal performance of CGAE when anomalous data from 1 or 2 runs is used is also lower than that of MCGAE, and that the difference between the performance using the optimal and chosen threshold is, in general, lower for CGAE than MCGAE.

\section{Relative difference in thresholds for the different reconstructions}\label{appendix:ths_reconstructions}

This appendix will discuss the (relative) difference between an "optimal" threshold and a chosen threshold, by \eqref{eq:threshold_CGAE}, for the ablation study with regards to different reconstructions for MCGAE.
The same metric as in Appendix \ref{appendix:ths_methods} will be used.

\subsection{Results}

\begin{figure}[ht]
	\centering{\includegraphics[keepaspectratio=false,width=\columnwidth]{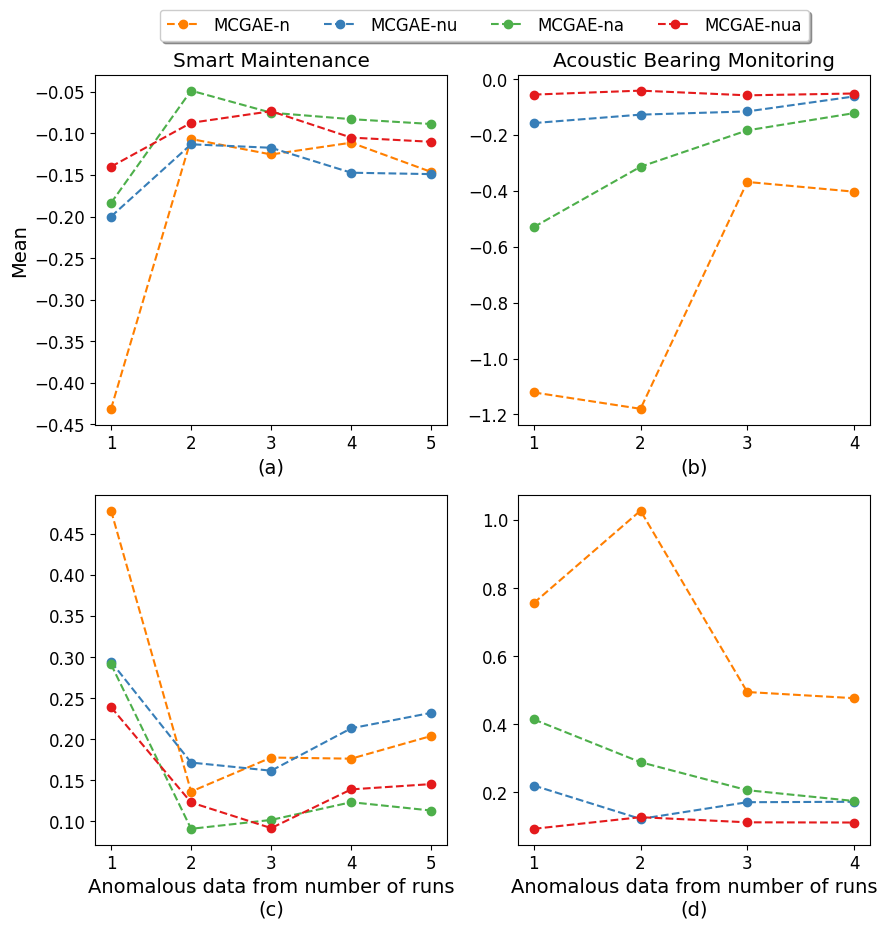}}
	\caption{The mean (a) and standard deviation (c) of the relative difference in norm between the different thresholds and the corresponding optimal threshold for the BA on the SM dataset, as well as the ABM dataset (b) and (d), for the different reconstructions.}
	\label{fig:Ablation_th}
\end{figure}

The results for the comparison between the thresholds based on training statistics and an optimal threshold for the different reconstructions are shown in Figure \ref{fig:Ablation_th}.
For the SM dataset, the largest difference between the reconstructions can be seen when using anomalous data from only 1 run, where MCGAE-n shows a noticeably more negative difference, indicating that the chosen threshold is significantly larger than the optimal threshold.
When anomalous data from more runs is used, the relative difference is similar to the other reconstructions.
Overall, MCGAE-na and MCGAE-nua do show a lower, and thus better, relative difference.
However, it should be noted that the difference with MCGAE-n and MCGAE-u is only small.
This behavior is similar to what was noticed when evaluating the discriminative performance.
For the ABM dataset, the relative difference varies strongly between the reconstructions, with MCGAE-n showing the largest negative difference between the chosen and optimal threshold.
As this indicates that the chosen threshold is significantly lower than the optimal, it is likely that the smaller differences for the other reconstructions are due overall higher estimated CI for the normal data causing the optimal threshold to increase.
As discussed when evaluating the discriminative performance of the reconstructions on the ABM dataset, it was noticed that the MCGAE-nu and MCGAE-nua show and overall higher estimated CI for the normal data, likely due to the reconstruction of unlabeled data, and that this was not as noticeable for MCGAE-na.
This is also represented in these results, as MCGAE-na shows a larger relative difference in comparison to MCGAE-nu and MCGAE-nua, which could be attributed to the smaller increase in the estimated CI for the normal data.

\subsubsection*{Acknowledgements}

\textbf{Funding} This research received funding from the Flemish Government (AI Research Program).
The work presented in this paper was supported by the ACMON ICON project and the CAATS ICON project.
This research has received support of Flanders Make, the strategic research centre for the manufacturing industry. \newline

\noindent\textbf{Author Contributions} Maarten Meire and Quinten van Baelen were responsible for the implementation, execution and analysis of the experiments and the writing of the manuscript. Ted Ooijevaar was responsible for the data and reviewing the paper. Peter Karsmakers was responsible for supervision and reviewing the paper. \newline

\noindent\textbf{Data availability} The datasets used in this paper were provided by Flanders Make. \newline

\noindent\textbf{Code availability} The code is available upon request to the authors.

\subsubsection*{Declarations}

\textbf{Conflict of Interest} The authors have no competing interests to declare that are relevant to the content of this article. \newline

\noindent\textbf{Ethics Approval} Not applicable. \newline

\noindent\textbf{Consent to participate} Not applicable. \newline

\noindent\textbf{Consent for publication} All authors consent for submission, and publication.

\bibliography{sn-article}

\end{document}